\documentclass[lettersize,journal, twoside]{IEEEtran}
\usepackage{graphicx}
\usepackage{booktabs} 
\usepackage{paralist}
\usepackage[inkscapelatex=false]{svg}
\usepackage[shortlabels]{enumitem}
\usepackage{hyperref}
\usepackage{multirow}
\usepackage{colortbl}
\usepackage{tabulary}
\usepackage{etoolbox}
\usepackage{makecell}
\usepackage{tikz}
\usepackage{changepage}
\usepackage{pifont}

\usepackage{amsmath}
\usepackage{mathptmx}

\usepackage{algorithm}
\usepackage[noend]{algpseudocode}
\usepackage{algcompatible}
\usepackage{lineno}
\usepackage{comment}
\usepackage{amssymb}
\usepackage{microtype}
\usepackage{bm}
\usepackage{colortbl}
\usepackage{cite}

\definecolor{control}{rgb}{0.1, 0.2, 0.5}
\definecolor{action}{rgb}{0.8, 0.1, 0.2}

\begin{document}

\title{A Human-in-the-loop Approach to Robot Action Replanning through LLM Common-Sense Reasoning}

\author{Elena Merlo, Marta Lagomarsino, and Arash Ajoudani
\thanks{This work was supported by the European Union Horizon Project TORNADO (GA 101189557). The authors are with Human-Robot Interfaces and Interaction (HRII) Laboratory, Istituto Italiano di Tecnologia, Genoa, Italy. Elena Merlo is also with the Dept. of Informatics, Bioengineering, Robotics, and Systems Engineering, University of Genoa, Genoa, Italy. Corresponding author's email: {\tt\footnotesize elena.merlo@iit.it}}
\vspace{-1.2cm}
}

\markboth{IEEE Robotics and Automation Letters. Preprint Version. Accepted MM, YYYY}%
{Merlo \MakeLowercase{\textit{et al.}}: A Human-in-the-loop Approach to Robot Action Replanning through LLM Common-Sense Reasoning}


\maketitle

\begin{abstract}
To facilitate the wider adoption of robotics, accessible programming tools are required for non-experts. Observational learning enables intuitive human skills transfer through hands-on demonstrations, but relying solely on visual input can be inefficient in terms of scalability and failure mitigation, especially when based on a single demonstration. This paper presents a human-in-the-loop method for enhancing the robot execution plan, automatically generated based on a single RGB video, with natural language input to a Large Language Model (LLM). By including user-specified goals or critical task aspects and exploiting the LLM common-sense reasoning, the system adjusts the vision-based plan to prevent potential failures and adapts it based on the received instructions. Experiments demonstrated the framework intuitiveness and effectiveness in correcting vision-derived errors and adapting plans without requiring additional demonstrations. Moreover, interactive plan refinement and hallucination corrections promoted system robustness. 
\end{abstract}

\begin{IEEEkeywords}
Human-in-the-loop; Interactive planning; Large Language Models (LLM); Failure mitigation
\end{IEEEkeywords}
\vspace{-0.4cm}
\section{Introduction}
\vspace{-0.1cm}
\IEEEPARstart{A}{s} robots become increasingly integrated into households and workplaces \cite{lorenzini2023ergonomic}, they have the potential to serve as versatile tools for a wide range of tasks. This shift means that more everyday users, with little to no experience in robotics or programming, will need to instruct robots for their specific needs. 
Programming by Demonstration (PbD) \cite{billard2008survey, ravichandar2020recent} facilitates the human skills transfer, allowing users to teach robots via hands-on demonstrations rather than traditional coding. This paradigm takes cues from human social learning processes such as emulation \cite{whiten2009emulation}.

Within PbD, observational learning \cite{pauly2021seeing} entails the systematic observation of a human demonstrator and their surroundings to identify relevant objects \cite{ramirezamaro2017transferring}, monitor environmental changes, and recognize actions along with their effects and preconditions \cite{zanchettin2023symbolic}. Through this process, the system can derive an execution plan for the robot that preserves the logical sequence of steps required to achieve a specific goal \cite{merlo2024exploiting, diehl2021automated} (e.g., placing a pen in a case, cutting cheese, or cleaning a kitchen), and replicate the demonstrated task by following analogous motion patterns \cite{arduengo2023gaussian}.
However, despite continuous advancements in computer vision and action recognition \cite{beddiar2020vision, jegham2020vision}, human and object motion tracking algorithms still make errors. This makes visual information often insufficient for automatically generating a reliable plan to replicate the execution, particularly when relying on a single demonstration, as recommendable in intuitive programming \cite{orendt2016robot}.

To address inaccuracies in visual data, researchers have combined them with natural language descriptions, implementing language-conditioned PbD methods \cite{stepputtis2020language, yu2022using, shao2021concept2robot, mees2022matters, jang2022bc, goyal2021zero}. These approaches are effective because they incorporate objects' spatio-temporal relationships and trajectory-level data derived from vision, while language clarifies any uncertain or missing contextual information and specifies high-level intentions or task goals that may not be apparent from the visual data alone. This helps in performing successful robot replicas of the learned task in different contexts, even when providing a limited number of demonstrations.

The emergence of pre-trained Large Language Models (LLMs) \cite{achiam2023gpt} has enhanced the potential of video-text integration in PbD. Their extensive training enables flexible, template-free language input and allows the generation of complex, enriched outputs that go beyond the provided inputs by leveraging the model internal knowledge. For instance, in \cite{wang2023demo2code, murray2024teaching}, the user teaches skills to the robot by providing both language descriptions and visual demonstrations to an LLM that generates structured and generalizable manipulation programs, incorporating high-level logic like loops and conditions. In \cite{wake2024gpt} the authors propose a multimodal pipeline that elaborates a video demonstration integrating user language feedback to generate task plans and extract key affordances for robot execution. Despite these advances, LLMs can still hallucinate, producing unfeasible or unsafe plans if not properly constrained \cite{wang2024large}. As an alternative, some works use LLMs not to generate plans from scratch but to refine existing ones. In \cite{sharma2022correcting, bucker2023latte}, the user provides corrective instructions, such as modifying goals, adding constraints, or specifying waypoints, which the LLM translates into adjustments to the robot’s motion plan. In \cite{shi2024yell}, real-time language corrections influence the selection and adjustment of low-level actions by guiding high-level policy decisions. While \cite{yu2023language} proposes using LLMs to generate reward functions, which are then optimized in real-time to bridge high-level language instructions and low-level robot actions for interactive task execution. In \cite{cui2023no}, the authors introduce a shared autonomy system that maps high-level language instructions and real-time verbal corrections into dynamic, low-dimensional joystick control spaces, allowing users to guide and refine a manipulator behavior during execution.
Although such approaches involve humans in the loop, the user's role is often limited to local corrections of robot motion during execution, without visibility or control over the full task plan, which they might want to review, modify, or personalize.

\begin{figure*}[t]
    \centering
    \vspace{-0.8cm}
    \includegraphics[width=0.95\linewidth]{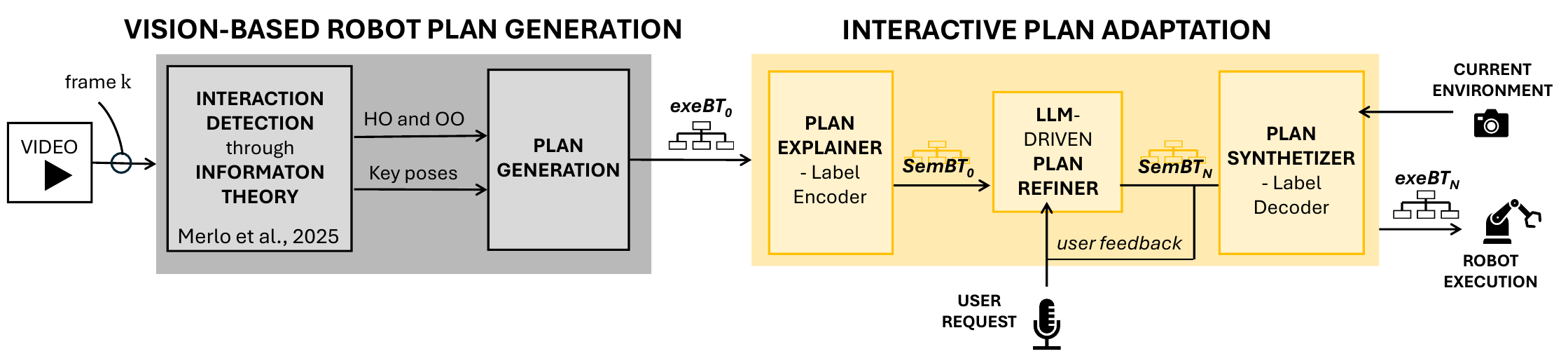}
    \vspace{-0.4cm}
    \caption{The system processes video demonstration frames to generate a behavior tree (gray block), converts it to a semantic version, and allows users to refine it with verbal instructions to the LLM. The LLM-enhanced plan is then translated back into an executable format (yellow block).}
    \label{fig:framework}
    \vspace{-0.6cm}
\end{figure*}

To address this challenge, we propose a pipeline that allows users to interact with and refine a complete execution plan, generated from a single RGB video demonstration of manual tasks, through natural language. 
Building on our previous work, the robot plan is generated as a Behavior Tree (BT) \cite{merlo2024exploiting}, as shown in the gray block in Fig. \ref{fig:framework}, using Shannon’s Information Theory (IT) to extract the task-relevant action sequence and identify key hand trajectory waypoints.
Since vision-based plans may contain pose estimation errors, users iteratively correct and personalize them by interacting with an LLM, which adjusts BT nodes and parameters based on their requests. This process allows users to maintain global supervision until they are satisfied and execute the plan.
In summary, the contribution of this paper is the design and implementation of a novel human-in-the-loop strategy that integrates LLM reasoning capabilities to refine one-shot, video-generated robot plans. This approach offers the following advantages:
\begin{inparaenum}[(i)]
\item vision-related errors can be intuitively corrected by non-expert users;
\item the initial plan can be adapted or extended just by furnishing new task requirements in natural language;
\item human supervision helps identify and address logical hallucinations produced by the LLM, promoting overall system robustness;
\item users retain global control over the entire task plan.
\end{inparaenum}
\vspace{-0.2cm}

\section{Methods}
\vspace{-0.1cm}
The pipeline consists of two modules as shown in Fig. \ref{fig:framework}. The first one (in gray) processes each frame of the video demonstration, analyzing hand-object and object-object interactions (HO and OO, respectively), using entropy measures from Shannon's IT. This analysis segments the executed task into Interaction Units (IUs), temporal blocks where scene interactions remain stable. By identifying the hand actions that trigger IU transitions, a set of robot action primitives is extracted and mapped into a BT plan. Note that semantics is never used in this procedure; we study the information content during interactions between video elements without assigning meaning to them. For further details, refer to \cite{merlo2024exploiting}.
The second module (main contribution of this paper, in yellow) enables user's interaction with the LLM to refine the generated BT and execute it once satisfied. It consists of three sub-blocks: the \emph{Plan Explainer}, which translates the vision-based plan \textit{exeBT$_0$} from numerical to semantic \textit{SemBT$_0$}, facilitating the user to review and identify necessary refinements; the \emph{Plan Refiner}, which handles user's requests and returns an updated plan version using LLM; and the \emph{Plan Synthetizer}, which reconverts the final plan \textit{SemBT$_N$} into its executable version \textit{exeBT$_N$} for the robot replica.
\vspace{-0.2 cm}

\subsection{Vision-based Robot Plan Generation}
This section revisits two aspects of \cite{merlo2024exploiting} to clarify the LLM-driven refinement mechanism of the generated plan. 
\subsubsection{Target Poses Extraction during Manipulation using Information Theory}
As mentioned earlier, the information flow between hands and objects in manipulation tasks helps segment hand activities that cause environmental changes. To determine if the hand and an object at frame $k$ of the video share information, we use Mutual Information (MI), which derives from IT and measures the dependency between two signals. 
We track the $3$D position of the hand \textbf{X} and the object \textbf{Y} and compute MI for each spatial dimension by pairing corresponding components ($MI(X^{(i)} : Y^{(i)})$, for $i = 1, 2, 3$). The $1$D positional signals are processed directly without filtering, which highlights one strength of using IT measures, shown to be more robust to noise than classical velocity-based approaches \cite{merlo2024exploiting}. The computation is done over a shifting time window $w$ centered at $t^*$, when frame $k$ was taken:  
\begin{equation}
\label{eq:mi}
MI(X^{(i)}(t^*):Y^{(i)}(t^*)) = \sum_{x \in \Omega_x} \sum_{y \in \Omega_y} p_{xy}(x,y) \cdot \log_{2}{\frac{p_{xy}(x, y)}{p_x(x)p_y(y)}},
\end{equation}
where $p_x(x)$ and $p_y(y)$ are the probabilities of $X^{(i)}$ and $Y^{(i)}$ taking values $x$ and $y$ within $w$, respectively, and $p_{xy}(x,y)$ is their joint probability. The final $MI(\textbf{X}(t^*):\textbf{Y}(t^*))$ value is the sum of the three component-wise MI. If $MI(t^*) = 0$, the hand and the object moved independently in $w$, otherwise they were correlated, thus in interaction.  
By computing MI over the hand and the in-hand object trajectories within shifting $w$ helps also detect motion pattern changes during manipulation. A constant $MI(t)$ indicates steady hand positional variability, while when the hand slows down and stops, its position values become more predictable, leading to a decreasing $MI(t)$.
For instance, when the temporal trend of $MI(t)$ forms a bell-shaped curve, it corresponds to a pick-and-place task where the grasping phase corresponds to the increasing slope, the transport phase to the steady peak, and the release phase to the decreasing slope.
When $MI(t)$ results in a more waved signal, each valley indicates that the hand has slowed down before accelerating again or has repeatedly assumed the same position within $w$, performing confined movements. For instance, in a back-and-forth motion during a cutting task, the signal takes on a wave-like pattern, where the valleys correspond to moments when the elementary movement (back or forth) is completed, the direction changes, and the hand and manipulated knife assume similar positions within a short time frame. These local minima in $MI(t)$ correspond to turning points, as the high probability of passing through these poses indicates their significance in the performed movement.
We use these key poses to generate a simpler yet effective trajectory for the robot. This eases the skill transfer compared to collecting data to enable the imitation of the full motion of the human hand. Note that the $MI(t)$ analysis simplifies the extraction of such key poses by relying on a single signal, avoiding the need to track multiple velocity components. 
At each minimum, the relative pose between the manipulated object $o_m$ (e.g., the knife) and the background object $o_{bkg}$ (e.g., the bread) is recorded. From now on, we refer to these relative object-object poses as Target Poses (TPs) and to the moments in which they occur as \textit{key instants}. Once given the current environment (i.e., object configurations), these TPs are transformed into waypoints for the robot's end-effector that let $o_m$ assuming the same set of relative poses with respect to $o_{bkg}$.
The reliance on relative poses makes the approach robust to changes in object layout, as long as the objects remain reachable by the end-effector.
One additional TP is extracted: the one at the first instant of the OO, representing the relative pose during the approaching phase. This new key instant is determined by considering when the 
average object-object distance $\overline{d}_{o_m, o_{bkg}}$ over window $w$ first falls below a threshold $d^{th}_{oo}$, indicating sufficient proximity (e.g., knife approaching bread).
\subsubsection{Automatically Generated Vision-based Plan}
A BT is a hierarchical structure used in robotics to model execution plans. The root node sends a tick signal to propagate through control nodes, which select a policy to execute actions or evaluate conditions. Each node returns a status among SUCCESS, FAILURE, or RUNNING to guide task execution.
In our previous work \cite{merlo2024exploiting}, the action nodes in our BT were of two types: \textit{Grasp}, which controls the opening and closing of the gripper, and \textit{ExecTrajectory}, which handles arm movement from a start to a target point. In \cite{merlo2024exploiting}, we also detail how we extract from the video demonstration the sequence of such robot actions to reach the observed goal. 
In this work, we extend the functionality of \textit{ExecTrajectory} to handle the arm movement through a sequence of TPs, rather than just a single final TP.
As a result, for tasks like cutting, the \textit{ExecTrajectory} node guides the gripper through each extracted TP step by step.
Each $\text{TP}_i$ is represented as a transformation matrix $T_{o_m}^{o_{bkg}}$ and provided as a value for the attribute \textit{target-pose$_i$} of the \textit{ExecTrajectory} node.

\vspace{-0.3 cm}
\subsection{Interactive Plan Adaptation}
This block processes the user's verbal commands and leverages the LLM common-sense reasoning to correct and adapt the vision-generated plan.

\subsubsection{Label Encoder - Plan Explainer}

This block is responsible for transforming the numerical vision-based \textit{exeBT$_0$} into its semantic version, \textit{SemBT$_0$}, by replacing numbers with human-interpretable descriptions. This process is particularly relevant for the \textit{target-pose$_i$} attributes of \textit{ExecTrajectory} node, which is translated from $T_{o_m}^{o_{bkg}}$ into a structured sentence in natural language, allowing human users to interpret and verify task execution steps. 
This sentence includes a triplet of labels that describe the pose of $o_m$ relative to $o_{bkg}$, considering (i) its position in the horizontal plane, (ii) its vertical displacement, and (iii) an estimate of its orientation.  
To obtain the first label, we consider that each object has a set of Interaction Points (IPs), representing key reference locations (e.g., corners of a box, tip of a pen) relative to the object’s centroid. To convert numerical poses into a semantic description, the position of $o_m$ is substituted with the name of the closest interaction point of $o_{bkg}$, identified considering the smallest Euclidean distance between $o_m$ centroid and all the $o_{bkg}$ IPs. This provides the first intuitive spatial reference. 
Then, the vertical displacement between $o_m$ centroid and the closest $o_{bkg}$ IP (namely along $z$ axis) $\delta_z=z_{o_m}-z_{o_{bkg}}^{IP}$ is separately analyzed and classified based on a threshold $z^{th}$ into three meaningful labels: \textit{above} if $\delta_z > z^{th}$, \textit{touching} if $-z^{th} \leq \delta_z \leq z^{th}$, and \textit{below} if $\delta_z < -z^{th}$. 
Finally, to describe $o_m$ orientation, we compare its orientation encoded in $T_{o_m, k}^{o_{bkg}}$, with that at the previous key instant, $T_{o_m, k-1}^{o_{bkg}}$.
We focus only on the most significant rotation in $[k-1, k]$ to ensure that the semantic description captures it. The rotation axis is denoted using the following terms: \textit{side bending} for rotations around the $x$-axis, \textit{tilting} for the $y$-axis, and \textit{turning} for the $z$-axis.
Then, the extracted rotation angle $\theta$ is mapped to the closest angle $\theta^*$ among predefined values:

$\theta^* = \underset{\theta'}{\operatorname{nearest}} \{0^\circ, 45^\circ, 90^\circ, 135^\circ, 180^\circ, -135^\circ, -90^\circ, -45^\circ\}.$
The combination of the identified rotation axis and $\theta^*$ defines the third target label. 
For example, if the task is positioning a cup on a plate, the final target pose could be encoded in the following sentence: \textit{plate center, touching, turning $90^\circ$}.

\subsubsection{LLM-driven Plan Refiner}
Once \textit{SemBT$_0$} plan is generated, it is possible to dialogue with the LLM, asking for modifications and improvements.
The LLM is initially provided with the \textit{SemBT$_0$} in XML format, IPs labels of the involved objects, structured refinement guidelines, and user requests. 
Its task is to enhance the \textit{SemBT$_0$} by correcting errors due to vision limitations, such as perception inaccuracies, based on user's feedback. 
The guidelines provide the LLM with a description of the XML structure and each node attribute. Specifically, the three-label format for the \textit{ExecTrajectory} \textit{target-pose$_i$} attribute is highlighted, implicitly asking the model to keep it. Moreover, for tasks it recognizes to involve contact forces, it is instructed to include the stiffness attribute in \textit{ExecTrajectory} and specify the arm stiffness level: \textit{high} for precise tracking, \textit{medium} (default) for general tasks, and \textit{low} for compliance-heavy tasks.
The LLM-refined XML is checked for structural correctness (e.g., node closures, tree integrity), and missing elements are automatically fixed to ensure compatibility with the execution pipeline.
Then, \textit{SemBT$_j$} (with $j= {1, \dots, N}$ denoting the refinement iteration) becomes available for user's review, allowing them to identify and address remaining logical hallucinations or incoherent adaptations. 
Thus, once the errors in the vision-based BT are addressed, the user can request further adjustments to the plan to meet their current needs while remaining in the loop. 
For instance, in the cup replacement task, the user could specify that ``the cup is full''. This additional detail may lead the model to increase the transportation time of the cup to reduce the probability of spilling.
User's commands can also induce the LLM to remove or add some new nodes, always keeping faith to the XML structural guidelines. 
The LLM operates with contextual input, combining the guidelines (static context) with dynamic context that includes the latest user instruction and the most recent semantic plan \textit{SemBT$_j$}. If the user is unsatisfied with the outcome, they can restore the previous version \textit{SemBT$_{j-1}$}, effectively discarding the last instruction and LLM output. This mechanism supports clearer rephrasing and helps the system converge to a valid response.


\subsubsection{Label Decoder - Plan Synthetizer}
This block is responsible for the inverse operation made by the label encoder. It takes the final \textit{SemBT$_N$} and prepares it for being executable (\textit{exeBT$_N$}). By parsing \textit{SemBT$_N$}, nodes that were already present in $SemBT_0$ obtained from vision data are reconverted to their numerical form, maintaining consistency with the vision-based \textit{exeBT}$_0$. When either an LLM-varied or -added \textit{target-pose$_i$} is recognized, the corresponding numerical version of the encoded TP is computed. In particular, the closest interaction point determines the $x$ and $y$ coordinates of $o_m$. The vertical displacement label sets the $z$ coordinate to $ z_{above}$ if it is $above$, to $ z_{below}$ if is $below$ or to $\epsilon_z$ in case of $touching$. Finally, the rotation label is converted into a rotation matrix around the indicated axis. These values define the matrix $T^{o_{bkg}}_{o_m}$ for the \textit{ExecTrajectory} node in \textit{exeBT$_N$}.

\begin{figure} [t]
    \centering
    \includegraphics[width=0.7\linewidth]{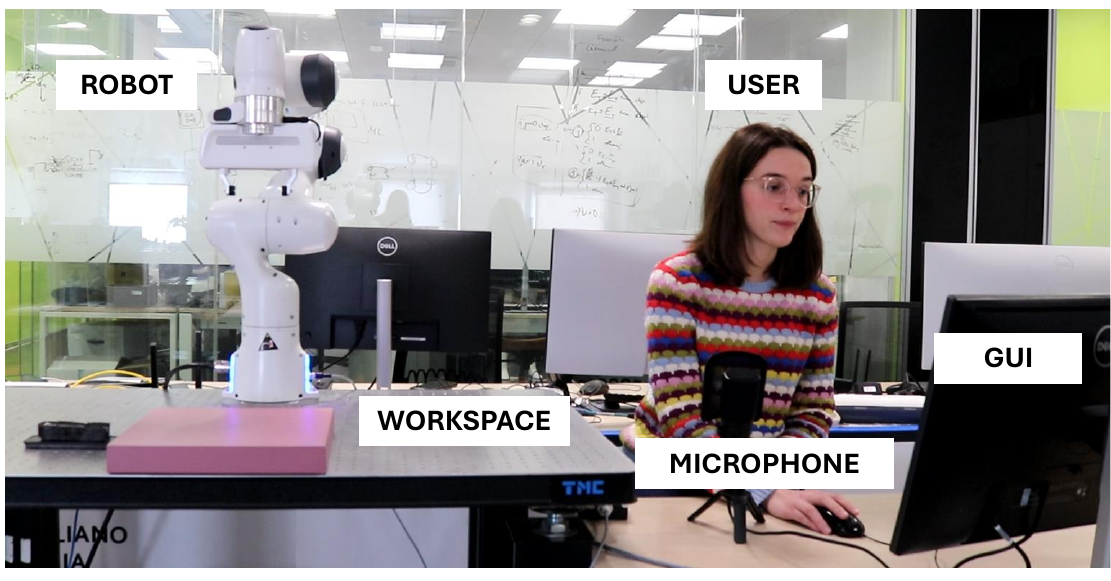}
    \vspace{-0.3cm}
    \caption{Experimental setup: the user consults the Graphical User Interface displaying the semantic robot plans and asks the LLM for refinements through the microphone. The robot is ready to manipulate objects in the workspace and will start moving once she sends the final plan. }
    \label{fig:expe-setup}
    \vspace{-0.6cm}
\end{figure}

\vspace{-0.3cm}\section{Experiments}
The experimental campaign assessed the performance of our method through three distinct experiments\footnotemark[1].
First, we conducted a pilot ablation study to demonstrate system functionality by comparing robot performance with and without the LLM-powered semantic abstraction layer. Secondly, a multi-subject experiment assessed the robustness and usability of the framework across users with varying levels of expertise in robot programming. 
Finally, the LLM reasoning was tested with high-level requests of varying complexity, and the resulting strategies were analyzed.

\vspace{-0.4cm}
\subsection{Validation of Interactive Refinement of Vision-Based Plans}
A pilot study was conducted in which a researcher interacted with the model to refine failed vision-based robot plans generated for two distinct human-demonstrated tasks: (i) manipulating a jug to pour water into a glass and (ii) using a sponge to clean the surface of a tray. The user engaged in the interactive loop once per task.
The difficulty of inferring the correct jug orientation from visual data and the need to manage interaction forces between the sponge and tray make these two tasks challenging. 
We used a marker-based system (AruCo) to detect the $6$D poses of hands and objects during demonstrations, attaching markers to the back of the hand and to objects to preserve natural manipulation.
Note that markerless object and hand detection methods are rapidly advancing \cite{wu2024general}, offering promising opportunities for integration into our framework. We defined the following IPs for the two $o_{bkg}$, namely glass and tray: IP$_{\text{glass}} = \{$\textit{left rim, right rim, center}$\}$, IP$_{\text{tray}} = \{$\textit{bottom-left corner, bottom-right corner, top-left corner, top-right corner, bottom-edge mid point, top-edge mid point, left-edge mid point, right-edge mid point, center}$\}$. The jug was reduced to a single point at its spout and the sponge to the central point of its front edge, where we had attached the marker.
To discriminate the vertical relationship between $o_m$ and $o_{bkg}$ we set $z^{th} = 0.01$ m. Instead, we chose $z_{above} = 0.15$ m and $z_{below} = -0.15$ m to ensure a clear separation between the objects and $\epsilon_z = 0$ m to ensure their contact.
\footnotetext[1]{A video illustrating the framework functioning can be found online at \href{https://youtu.be/vUvFn1GJfR0}{https://youtu.be/vUvFn1GJfR0}.}
The robot arm translation stiffness values were set to 1000 N/m, 1500 N/m, and 2000 N/m for \textit{low}, \textit{medium}, and \textit{high} levels, respectively.
Thanks to a Graphical User Interface (GUI), the subject could check the video demonstration and consult the automatically generated plan translated into its semantic version (see the experimental setup in Fig. \ref{fig:expe-setup}), visualized using Groot2 software \footnote{Visit \href{https://www.behaviortree.dev/groot/}{https://www.behaviortree.dev/groot/} for details about Groot2.}. Using the microphone she could ask for some changes in the plan, check again the updated plan and remain in the loop until the plan was satisfactory enough to be sent for robot execution. The LLM model we employed was GPT-4o from OpenAI \cite{openai2024gpt4o}, while for converting audio requests to LLM prompts we used Whisper, the automatic speech recognition model by OpenAI \cite{radford2023robust}.
Given the current pose of the objects, the \textit{exeBT$_N$} was computed, and we compared the robot’s performance when following the vision-based plan versus the LLM-refined plan.

\vspace{-0.2cm}
\subsection{Assessment of Framework Robustness and Usability}
The multi-subject experiment assessed the framework’s robustness and usability across different users. We asked $10$ subjects ($7$ men and $3$ women, with an average of $32.6$ years) to use our system to modify the vision-based plan of the cleaning task\footnote{Experiments were conducted at the HRII Lab, Istituto Italiano di Tecnologia (IIT), in compliance with the Declaration of Helsinki. The protocol received approval from the ethics committee of Azienda Sanitaria Locale (ASL) Genovese N.3 under Protocol IIT\_HRII\_ERGOLEAN 156/2020.}. Among the participants, subjects $4$, $7$, $8$, and $9$ had no prior experience with robot programming. First, we proposed a familiarization task to help users understand the request-to-BT generation loop. 
Then, starting from the same BT, generated using a ``Z''-shaped cleaning motion on the top part of the tray, participants were asked to refine it to clean the entire surface. This was the only instruction; the refinement strategy was up to them.
At each iteration, users rated whether the LLM plan modifications matched their requests by selecting: Satisfied, Quite Satisfied, or Not Satisfied.
Once completed the task, participants evaluated their experience through the System Usability Scale (SUS) questionnaire, rating $10$ questions from $1$ to $5$. The experimenter also reviewed all request-BT pairs to identify potential causes of users' dissatisfaction. The average number of refinement requests was also recorded.

\vspace{-0.2cm}
\subsection{Evaluation of LLM Reasoning in Plan Adaptation}  
In the third experiment, we tested LLM common-sense reasoning by prompting it with $20$ GPT-4o-generated high-level requests to adjust a correct pouring plan, $10$ to pour less and $10$ to pour more.
We classified requests as medium complexity if they included explicitly \textit{pour} with \textit{less} or \textit{more} (or synonyms), and high complexity if the command was implicit (see Table \ref{tab:expe3}). The experimenter assessed whether each request was fulfilled and identified the strategy used by the LLM.

\vspace{-0.1cm}
\section{Results}
\subsection{Validation of Interactive Refinement of Vision-Based Plans}
\subsubsection*{Pouring Task} 
\begin{figure}
    \vspace{-0.2cm}
    \centering
    \includegraphics[width=0.67\linewidth]{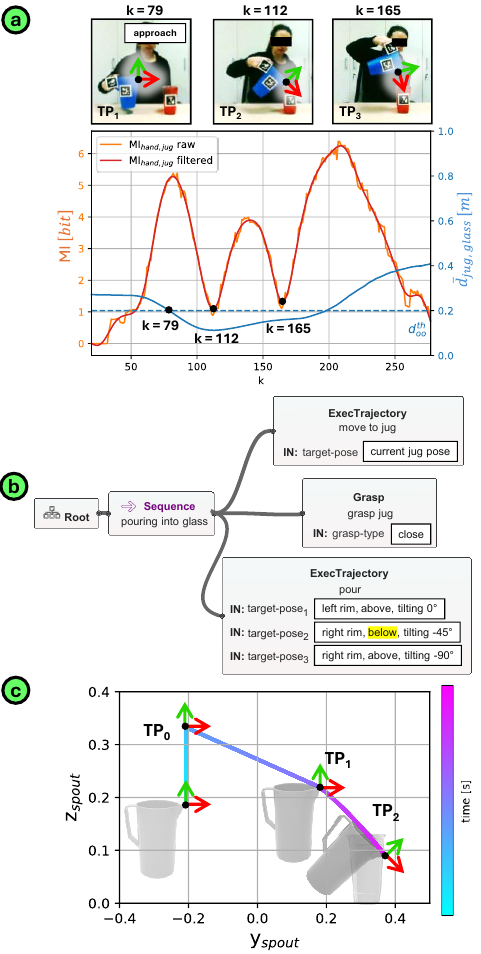}
    \vspace{-0.3cm}
    \caption{(a) Hand-jug Mutual Information signal (orange/red) and jug-glass distance (blue) during human pouring task; the retrieved key instants (black dots) during the approach and the jug-glass active interaction ($MI(t)$ minima); frames depicting the corresponding jug-glass target poses; (b) vision-based plan $exeBT_0$ in its semantic version \textit{SemBT}$_0$; (c) jug trajectory executing $exeBT_0$. }
    \label{fig:pouring-v}
    \vspace{-0.7cm}
\end{figure}

\begin{figure}
    \centering
    \vspace{-0.7cm}    
    \includegraphics[width=0.66\linewidth]{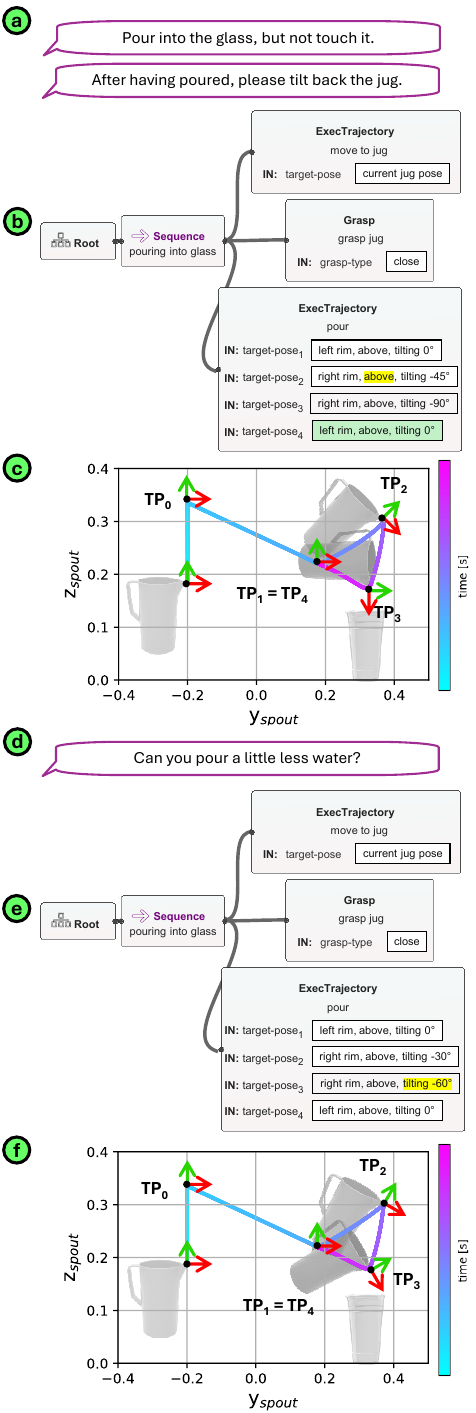}
    \vspace{-0.4cm}
    \caption{(a) User's requests to correct the vision-based plan; (b) LLM-generated \textit{SemBT} following user's input; (c) jug trajectory executing the LLM-enhanced BT; (d) user's request to adapt the plan for smaller water amount; (e) \textit{SemBT} generated to encounter this input; (f) jug trajectory executing the new BT.}
    \label{fig:pouring-v-llm}
    \vspace{-0.7cm}
\end{figure}

Fig. \ref{fig:pouring-v}-a presents the extraction of jug-glass TPs occurring during human pouring. The approaching key instant was identified when $\overline{d}_{jug,glass} < d_{oo}^{th}$ (blue line) at $k=79$. The minima of the filtered $MI_{hand,jug}$ signal (red line) at $k=112$ and $k=165$ indicate two distinct steps in rotating the jug while pouring. In the resulting \textit{SemBT}$_0$ plan, reported in Fig. \ref{fig:pouring-v}-b, the \textit{ExecTrajectory} node contains three entries, one per extracted TP. By executing this only-vision-based plan, the robot failed. In Fig. \ref{fig:pouring-v}-c, the spout trajectory is depicted as a colored curve from cyan to magenta, indicating time evolution, with frames highlighting its orientation at the TPs. 
TP$_0$ is the default pose used to lift the object after grasping. TP$_1$ corresponds to the vision-detected jug approach pose, positioning the spout above the left rim of the glass while keeping the jug upright. However, when moving toward TP$_2$, the jug collided with the glass, interrupting the execution. The reason is that the spout was incorrectly detected as being \textit{below} the right rim during the first pouring step.
To correct the plan, the user made two iterative requests: first, to pour without touching the glass, and second, to tilt the jug back once the pouring was completed (see Fig. \ref{fig:pouring-v-llm}-a). The LLM modified the plan as shown in Fig. \ref{fig:pouring-v-llm}-b, changing the label for the vertical relation from \textit{below} to \textit{above}, and adding TP$_4$, re-proposing the approaching pose.
The adapted TPs in the LLM-refined BT were automatically converted into the end-effector waypoints and defined a successful spout trajectory (see Fig. \ref{fig:pouring-v-llm}-c). Notably, upon reaching the adapted TP$_2$, the jug remained closer to the right rim and tilted at $-45^\circ$ relative to the glass. However, the height has been adjusted from the previously perceived value, obtaining $z_{spout} = z_{glass}^{\textit{right rim}} + z_{above}$ m, to prevent collisions. The robot concluded its movement by returning the jug to TP$_4$, coinciding with TP$_1$.
Finally, the user requested an adaptation of the plan to pour less water (Fig. \ref{fig:pouring-v-llm}-d). In response, the LLM common-sense reasoning adjusted the BT parameters (Fig. \ref{fig:pouring-v-llm}-e) reducing the jug tilting angle at TP$_3$ from $-90^\circ$ to $-60^\circ$ to limit the water flow into the glass. The new angle value was decided by the LLM.
The resulting spout trajectory is depicted in Fig. \ref{fig:pouring-v-llm}-f.

\subsubsection*{Cleaning Task} 
\begin{figure}
    \centering
    \vspace{-0.7cm}\includegraphics[width=0.80\linewidth]{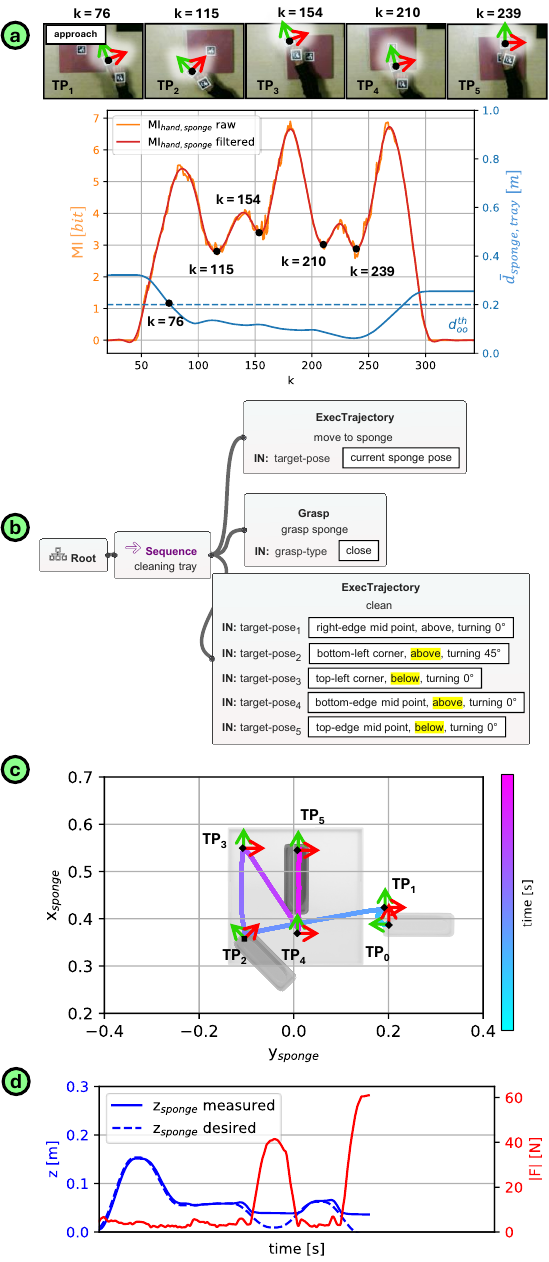}
    \vspace{-0.3cm}
    \caption{(a) Hand-sponge MI signal and sponge-tray distance during human cleaning task; retrieved key instants during the approach and the active sponge-tray interaction ($MI(t)$ minima); frames depicting the corresponding sponge-tray TPs; (b) vision-based plan $exeBT_0$ in its semantic version \textit{SemBT}$_0$; (c) sponge trajectory executing $exeBT_0$; (d) desired and measured $z_{sponge}$ during manipulation (blue) and magnitude $|\mathbf{F}|$ of the force exerted on the tray (red). }
    \label{fig:cleaning-v}
    \vspace{-0.7cm}
\end{figure}

In Fig. \ref{fig:cleaning-v}-a, the captured sponge-tray TPs occurring during human cleaning are presented. The sponge approached the tray at its right-edge mid point at $k=76$ when $\overline{d}_{sponge,tray}<d^{th}_{oo}$. $MI_{hand,sponge}$ exhibits a wavy pattern, with each valley corresponding to moments when the sponge slowed down near a specific region of the tray, often due to a change in direction. Such cleaning covered the left part of the tray following a zig-zag motion from the bottom-left corner to the top-edge mid point. This execution was converted in the $SemBT_0$ plan shown in Fig. \ref{fig:cleaning-v}-b, where the \textit{ExecTrajectory} node contains five entries, one per extracted TP. Note that all the TPs following the approach pose TP$_1$ have vertical relation labels different from the expected \textit{touching}, again for perception issues. The robot failed as the sponge did not maintain continuous contact with the tray, hindering the cleaning objective. 
In Fig. \ref{fig:cleaning-v}-d, the desired and measured $z_{sponge}$ (dashed and full blue line, respectively) during cleaning are shown with respect to the robot base. The first valley of the desired $z_{sponge}$ corresponds to the \textit{below} label at TP$_3$. In this case, the robot end-effector failed to follow the reference due to the presence of the tray, which is approximately $0.04$ m in height. This valley is followed by a peak caused by the \textit{above} label; the robot lifted the sponge, temporarily losing contact with the tray surface. A second, deeper valley appears, again associated with a \textit{below} label, but this time with a more negative value, causing the end-effector to press the sponge against the tray, exerting a force \textbf{F} with measured $|\textbf{F}| > 60$ N (red curve). In this execution, the robot arm stiffness was set to the standard (\textit{medium}) level.

\begin{figure} [t]
    \centering
    \vspace{-0.4cm}
    \includegraphics[width=0.80\linewidth]{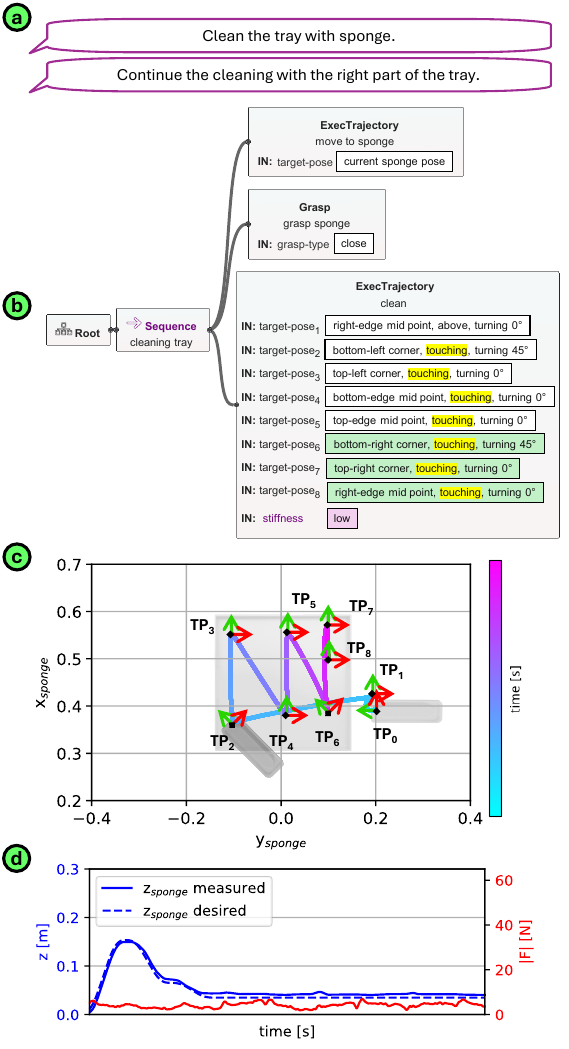}
    \vspace{-0.3cm}
    \caption{(a) User's requests to modify the vision-based plan; (b) LLM-generated \textit{SemBT} following user's input; (c) sponge trajectory executing the LLM-enhanced BT; (d) desired and measured $z_{sponge}$ during manipulation and magnitude $|\mathbf{F}|$ of the force exerted on the tray.}
    \label{fig:cleaning-v-llm}
    \vspace{-0.6cm}
\end{figure}

To correct the plan, the user first clarified that the task involved cleaning a tray with a sponge, helping the LLM infer the need for continuous contact between them.
As a result, all wrong z-relation labels were fixed. Additionally, recognizing that the task involved contact, the LLM set the \textit{stiffness} attribute of the \textit{ExecTrajectory} node to \textit{low} level (see Fig. \ref{fig:cleaning-v-llm}-b).
In a second iteration, the user requested to clean the right side of the tray as well. The LLM responded by generating three additional TPs, extending the zig-zag motion to cover the right area, as shown in Fig. \ref{fig:cleaning-v-llm}-c. Please note that the sponge poses TP$_1$ to TP$_5$ were replicated as they were detected, except for a modification in the $z$ coordinate. 
In Fig. \ref{fig:cleaning-v-llm}-d, the measured $z_{sponge}$ remained close to $0.04$ m, meaning sponge continuous contact with the tray surface, while the interaction force magnitude $|\textbf{F}|$ was lower thanks to the correction of the vertical displacements at all TPs and for the reduced stiffness.

\vspace{-0.2cm}
\subsection{Assessment of Framework Robustness and Usability}
\begin{figure}
    \centering
    \includegraphics[width=0.9\linewidth]{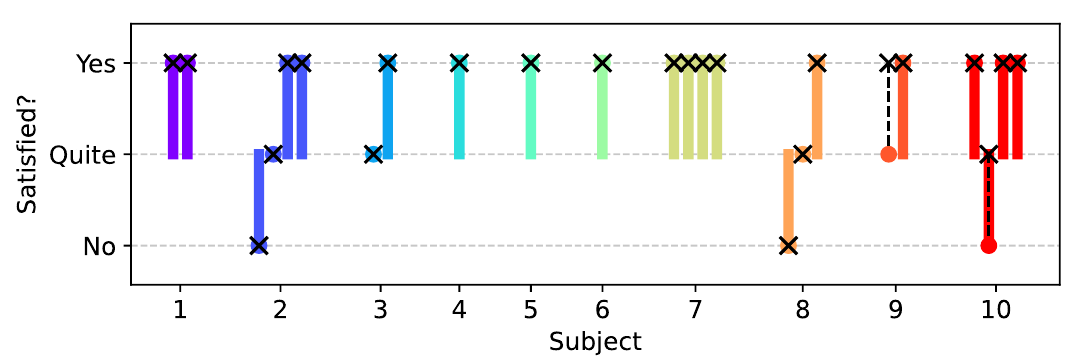}
    \vspace{-0.3cm}
    \caption{User's satisfaction with LLM-adapted plans across $10$ subjects. Each bar corresponds to a user's request, and the number of same-colored bars indicates the iterations required for the subject to obtain their desired version of the robot plan. Black crosses indicate the experimenter's satisfaction with each request-BT pair.}
    \label{fig:user-eval}
    \vspace{-0.4cm}
\end{figure}

Fig. \ref{fig:user-eval} reports user satisfaction with the LLM-empowered adaptations of the robot plans based on their requests. Each bar corresponds to a user request, and the number of bars with the same color indicates how many iterations the subject required to obtain their desired version of the robot plan. 
Users employed different strategies to complete the task, making diverse requests (detailed in appendix to experiment B \nameref{appendix}). 
Subjects $4$, $5$, and $6$ instructed the model with a single, precise command, mentioning that the task was a cleaning and specifying exactly the IP$_{\text{tray}}$ to touch with the sponge and in which order. 
Subjects $3$ and $9$ gave similar details about task semantics and specified the TPs but in two iterations. 
Subjects $7$ and $10$ preferred to ask to add a TP per iteration.
Subjects $1$ and $8$ obtained a comprehensive plan with more high-level requests: ``clean also the bottom part'' and ``complete the cleaning'' of the tray.

In response to these varied requests, the framework applied slightly different adaptation strategies. Importantly, all the subjects finally obtained a logical and executable plan that enabled the robot to complete the cleaning task successfully, with an average of $\overline{N} = 2.4$ iterations. 
Subjective user feedback revealed that out of $24$ total requests, the framework's adaptations were rated as only partially satisfying in $4$ instances, either because the LLM incorporated some but not all of the desired changes in the plan (as for subject $8$) or because it added modifications not explicitly requested (as for subjects $2$ and $3$). 
%
Logical hallucinations or incoherent adaptations were identified by users, as reflected by the $3$ unsatisfactory ratings of the LLM output. In these cases, users exploited the possibility to restore the previous BT and were able to refine the plan in the subsequent iterations to achieve their intended outcome. 

The black crosses represent the experimenter’s satisfaction with each request-\textit{SemBT} pair. Their evaluations align with user satisfaction in most cases, with a few exceptions. For subject $9$'s first command, the experimenter rated the LLM response as satisfactory, as the generated BT logically matched the request. The user's partial satisfaction was likely due to the absence of two TPs necessary to complete the cleaning, which, however, were not specified in the request. 
On the other hand, for the second request by subject $10$, we considered it partially satisfied since only some instructions were not addressed.

\begin{figure}
    \centering
    \includegraphics[width=0.75\linewidth]{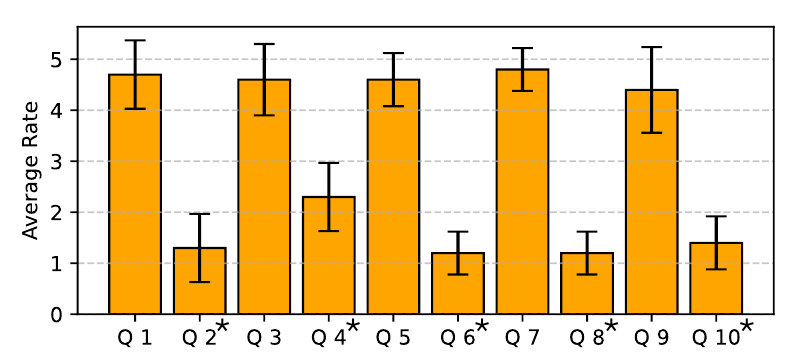}
    \vspace{-0.3cm}
    \caption{Average users' ratings for SUS questions represented by orange columns with error bars indicating variability. Higher bars indicate positive usability perceptions. Questions Qi* represent negative aspects, thus lower bars suggest fewer usability issues.}
    \label{fig:sus}
    \vspace{-0.5cm}
\end{figure}

In Fig. \ref{fig:sus}, we present the average scores assigned by users to the SUS questions. Users, especially those naive to robot programming, enjoyed using the system and found it easy to use and learn, with high ratings for Q1, Q3, and Q7. They also appreciated the integration of the system functionalities (Q5). Negative aspects like complexity, inconsistency, and cumbersomeness were rated low, indicating minimal usability issues. However, Q4 (need for support) and Q10 (learning effort) scored slightly higher, suggesting that some users may require assistance initially.

\begin{table}[t]
    \centering
    \caption{\small LLM Common-Sense Reasoning Evaluation}
    \vspace{-0.3cm}
    \begin{tabular}{cc c c}
        \toprule
        \toprule
        \multicolumn{2}{c}{\scriptsize Request} & & \scriptsize Sat \\
        \midrule
        \multirow{5}{*}{\rotatebox{90}{\scriptsize medium}} 
            & \scriptsize R1 & \scriptsize{Could you pour a bit less water into my glass, please?} & \textcolor{green!70!black}{\ding{51}}  \\
            & \scriptsize R2 & \scriptsize{I’d appreciate it if you could pour a little less.} & \textcolor{green!70!black}{\ding{51}}  \\
            & \scriptsize R3 & \scriptsize{Do you think you could give me a smaller amount of water?} & \textcolor{green!70!black}{\ding{51}}  \\
            & \scriptsize R4 & \scriptsize{Would you mind pouring just a little bit less water?} & \textcolor{green!70!black}{\ding{51}}  \\
            & \scriptsize R5 & \scriptsize{Can you pour just a little less, so it’s not too full?}  & \textcolor{green!70!black}{\ding{51}}  \\
        \midrule
        \multirow{5}{*}{\rotatebox{90}{\scriptsize high}} 
            & \scriptsize R6 & \scriptsize{Would you mind not filling the glass all the way up?} & \textcolor{green!70!black}{\ding{51}} \\
            & \scriptsize R7 & \scriptsize{Would you mind filling the glass a little less full?} & \textcolor{green!70!black}{\ding{51}} \\
            & \scriptsize R8 & \scriptsize{Can you leave some room in the glass for me?} & \textcolor{green!70!black}{\ding{51}} \\
            & \scriptsize R9 & \scriptsize{Could you stop before the glass gets too full?} & \textcolor{red!70}{\ding{55}} \\
            & \scriptsize R10 & \scriptsize{I think you are pouring too much water into the glass.} & \textcolor{green!70!black}{\ding{51}} \\
        \midrule
        \multirow{5}{*}{\rotatebox{90}{\scriptsize medium}} 
            & \scriptsize R11 & \scriptsize{Please pour more water into the glass.} & \textcolor{green!70!black}{\ding{51}}  \\
            & \scriptsize R12 & \scriptsize{Add more water to the glass.} & \textcolor{green!70!black}{\ding{51}}  \\
            & \scriptsize R13 & \scriptsize{Fill the glass with more water.} & \textcolor{green!70!black}{\ding{51}}  \\
            & \scriptsize R14 & \scriptsize{Give me more water in this glass.} & \textcolor{green!70!black}{\ding{51}}  \\
            & \scriptsize R15 & \scriptsize{Dispense a lot of water into the glass.}  & \textcolor{green!70!black}{\ding{51}}  \\
        \midrule
        \multirow{5}{*}{\rotatebox{90}{\scriptsize high}} 
            & \scriptsize R16 & \scriptsize{The glass should be filled.} & \textcolor{green!70!black}{\ding{51}} \\
            & \scriptsize R17 & \scriptsize{Keep pouring water until the glass is fuller.} & \textcolor{green!70!black}{\ding{51}} \\
            & \scriptsize R18 & \scriptsize{This glass is tall and needs more water to be filled.} & \textcolor{green!70!black}{\ding{51}} \\
            & \scriptsize R19 & \scriptsize{Empty the jug pouring into the glass.} & \textcolor{red!70}{\ding{55}} \\
            & \scriptsize R20 & \scriptsize{Could you leave a little less space in the glass?} & \textcolor{red!70}{\ding{55}} \\
        \bottomrule
    \end{tabular}
    \label{tab:expe3}
    \vspace{-0.5cm}
\end{table}

\vspace{-0.3cm}
\subsection{Evaluation of LLM Reasoning in Plan Adaptation} 
In Table \ref{tab:expe3}, each LLM-provided request (R) for adjusting the amount of water to pour into the glass is coupled with a tick (\textcolor{green!70!black}{\ding{51}}) or a cross (\textcolor{red!70}{\ding{55}}), indicating whether the experimenter evaluated the adaptation of the plan as satisfactory or not. The strategies adopted to reduce the water flow involved
\begin{inparaenum} [(i)]
    \item decreasing the pouring tilt angles (R1, R2, R3, R5, R7, R8),
    \item removing the pouring step with the greatest jug inclination (R6),
    \item or shortening the pouring duration by adjusting the execution time (R4, R10).
\end{inparaenum} 
Note that when reducing the angles or time, the proposed numerical values vary across requests. To respond to R9, the LLM modified only the tilt-back phase by increasing its execution time, which is illogical.
The strategies adopted to increase the poured water were
\begin{inparaenum} [(i)]
    \item inserting an additional pouring step after the tilt-back (R11),
    \item repeating the same pouring sequence twice (R12, R13, R14),
    \item adding more pouring steps, increasing the jug inclination before tilting it back (R15, R16, R17, R18).
\end{inparaenum}
The LLM failed to handle R19 and R20. In the latter case, it reduced the pouring angles, likely due to confusion caused by the presence of the word \textit{less}. 
However, these are hypotheses, as the LLM decision-making process is difficult to interpret due to its black-box nature.

\vspace{-0.3cm}
\section{Discussion and Conclusion}
We introduced a novel human-in-the-loop framework that enables non-expert users to refine high-level robot plans, generated from a single video demonstration, via natural language interaction with an LLM. Unlike prior approaches focused on low-level motion correction during execution, our method supports iterative, pre-execution refinement of interpretable BTs, granting users global control over the entire task plan that they can correct and customize.
This work demonstrates that LLM reasoning at a semantic level helps refine numerical data extracted from vision, such as trajectory waypoints, while also suggesting necessary modifications to non-visible parameters like arm stiffness. 
A current limitation is that the supported BTs consist of basic action sequences and lack condition checks (e.g., object presence) and parallelism, which would enable multi-agent execution. They also rely on a minimal set of action nodes that do not support more complex manipulations such as pushing or pulling, which would require dedicated nodes and parameters. Yet, future work will expand BT capabilities by also introducing semantic mappings for additional low-level parameters, enabling the LLM to reason over richer physical interactions. However, an increasing BT complexity raises the challenge of maintaining user interpretability and non-expert control over the plan. 

To support this, system guidelines will need to be updated to ensure correct LLM interpretation and refinements, while automatic handling of structural BT hallucinations that users cannot fix and tools such as a simulator or digital twin for plan verification could be investigated.
Moreover, we are exploring how video data, combined with pre-trained and fine-tuned LLM reasoning, can help infer appropriate robot control policies: position control for precision tasks (e.g., inserting a peg into a hole), impedance control for stable contact with stiff environments (e.g., wiping a rigid surface), and admittance control for safe co-manipulation, as soon as human-human collaboration videos become processable to generate human-robot collaboration plans.

\vspace{-0.3cm}
\bibliographystyle{IEEEtran}
\bibliography{biblio}

\begin{thebibliography}{10}
\providecommand{\url}[1]{#1}
\csname url@samestyle\endcsname
\providecommand{\newblock}{\relax}
\providecommand{\bibinfo}[2]{#2}
\providecommand{\BIBentrySTDinterwordspacing}{\spaceskip=0pt\relax}
\providecommand{\BIBentryALTinterwordstretchfactor}{4}
\providecommand{\BIBentryALTinterwordspacing}{\spaceskip=\fontdimen2\font plus
\BIBentryALTinterwordstretchfactor\fontdimen3\font minus \fontdimen4\font\relax}
\providecommand{\BIBforeignlanguage}[2]{{%
\expandafter\ifx\csname l@#1\endcsname\relax
\typeout{** WARNING: IEEEtran.bst: No hyphenation pattern has been}%
\typeout{** loaded for the language `#1'. Using the pattern for}%
\typeout{** the default language instead.}%
\else
\language=\csname l@#1\endcsname
\fi
#2}}
\providecommand{\BIBdecl}{\relax}
\BIBdecl

\bibitem{lorenzini2023ergonomic}
M.~Lorenzini, M.~Lagomarsino, L.~Fortini, S.~Gholami, and A.~Ajoudani, ``Ergonomic human-robot collaboration in industry: A review,'' \emph{Frontiers in Robotics and AI}, vol.~9, p. 813907, 2023.

\bibitem{billard2008survey}
A.~Billard, S.~Calinon, R.~Dillmann, and S.~Schaal, ``Survey: Robot programming by demonstration,'' \emph{Springer handbook of robotics}, pp. 1371--1394, 2008.

\bibitem{ravichandar2020recent}
H.~Ravichandar, A.~S. Polydoros, S.~Chernova, and A.~Billard, ``Recent advances in robot learning from demonstration,'' \emph{Annual review of control, robotics, and autonomous systems}, vol.~3, pp. 297--330, 2020.

\bibitem{whiten2009emulation}
A.~Whiten, N.~McGuigan, S.~Marshall-Pescini, and L.~M. Hopper, ``Emulation, imitation, over-imitation and the scope of culture for child and chimpanzee,'' \emph{Philosophical Trans. of the Royal Society B: Biological Sciences}, vol. 364, pp. 2417--2428, 2009.

\bibitem{pauly2021seeing}
L.~Pauly, ``Seeing to learn: Observational learning of robotic manipulation tasks,'' Ph.D. dissertation, University of Leeds, 2021.

\bibitem{ramirezamaro2017transferring}
K.~Ramirez-Amaro, M.~Beetz, and G.~Cheng, ``{Transferring skills to humanoid robots by extracting semantic representations from observations of human activities},'' \emph{Artificial Intelligence}, vol. 247, pp. 95--118, 6 2017.

\bibitem{zanchettin2023symbolic}
A.~M. Zanchettin, ``Symbolic representation of what robots are taught in one demonstration,'' \emph{Robotics and Autonomous Systems}, vol. 166, p. 104452, 2023.

\bibitem{merlo2024exploiting}
E.~Merlo, M.~Lagomarsino, E.~Lamon, and A.~Ajoudani, ``Exploiting information theory for intuitive robot programming of manual activities,'' \emph{IEEE Trans. on Robotics}, pp. 1--17, 2025.

\bibitem{diehl2021automated}
M.~Diehl, C.~Paxton, and K.~Ramirez-Amaro, ``{Automated Generation of Robotic Planning Domains from Observations},'' \emph{IEEE Intl. Conf. on Intelligent Robots and Systems}, pp. 6732--6738, 2021.

\bibitem{arduengo2023gaussian}
M.~Arduengo, A.~Colom{\'e}, J.~Lobo-Prat, L.~Sentis, and C.~Torras, ``Gaussian-process-based robot learning from demonstration,'' \emph{Journal of Ambient Intelligence and Humanized Computing}, pp. 1--14, 2023.

\bibitem{beddiar2020vision}
D.~R. Beddiar, B.~Nini, M.~Sabokrou, and A.~Hadid, ``Vision-based human activity recognition: a survey,'' \emph{Multimedia Tools and Applications}, vol.~79, pp. 30\,509--30\,555, 2020.

\bibitem{jegham2020vision}
I.~Jegham, A.~B. Khalifa, I.~Alouani, and M.~A. Mahjoub, ``Vision-based human action recognition: An overview and real world challenges,'' \emph{Forensic Science Intl.: Digital Investigation}, p. 200901, 2020.

\bibitem{orendt2016robot}
E.~M. Orendt, M.~Fichtner, and D.~Henrich, ``Robot programming by non-experts: Intuitiveness and robustness of one-shot robot programming,'' in \emph{2016 25th IEEE Intl. Symposium on Robot and Human Interactive Communication (RO-MAN)}.\hskip 1em plus 0.5em minus 0.4em\relax IEEE, 2016, pp. 192--199.

\bibitem{stepputtis2020language}
S.~Stepputtis, J.~Campbell, M.~Phielipp, S.~Lee, C.~Baral, and H.~Ben~Amor, ``Language-conditioned imitation learning for robot manipulation tasks,'' \emph{Advances in Neural Information Processing Systems}, vol.~33, pp. 13\,139--13\,150, 2020.

\bibitem{yu2022using}
A.~Yu and R.~J. Mooney, ``Using both demonstrations and language instructions to efficiently learn robotic tasks,'' \emph{arXiv preprint arXiv:2210.04476}, 2022.

\bibitem{shao2021concept2robot}
L.~Shao, T.~Migimatsu, Q.~Zhang, K.~Yang, and J.~Bohg, ``Concept2robot: Learning manipulation concepts from instructions and human demonstrations,'' \emph{The Intl. Journal of Robotics Research}, vol.~40, pp. 1419--1434, 2021.

\bibitem{mees2022matters}
O.~Mees, L.~Hermann, and W.~Burgard, ``What matters in language conditioned robotic imitation learning over unstructured data,'' \emph{IEEE Robotics and Automation Letters}, vol.~7, pp. 11\,205--11\,212, 2022.

\bibitem{jang2022bc}
E.~Jang, A.~Irpan, M.~Khansari, D.~Kappler, F.~Ebert, C.~Lynch, S.~Levine, and C.~Finn, ``Bc-z: Zero-shot task generalization with robotic imitation learning,'' in \emph{Conf. on Robot Learning}, 2022, pp. 991--1002.

\bibitem{goyal2021zero}
P.~Goyal, R.~J. Mooney, and S.~Niekum, ``Zero-shot task adaptation using natural language,'' \emph{arXiv preprint arXiv:2106.02972}, 2021.

\bibitem{achiam2023gpt}
J.~Achiam, S.~Adler, S.~Agarwal, L.~Ahmad, I.~Akkaya, F.~L. Aleman, D.~Almeida, J.~Altenschmidt, S.~Altman, S.~Anadkat \emph{et~al.}, ``Gpt-4 technical report,'' \emph{arXiv preprint arXiv:2303.08774}, 2023.

\bibitem{wang2023demo2code}
Y.~Wang, G.~Gonzalez-Pumariega, Y.~Sharma, and S.~Choudhury, ``Demo2code: From summarizing demonstrations to synthesizing code via extended chain-of-thought,'' \emph{Advances in Neural Information Processing Systems}, vol.~36, pp. 14\,848--14\,956, 2023.

\bibitem{murray2024teaching}
M.~Murray, A.~Gupta, and M.~Cakmak, ``Teaching robots with show and tell: Using foundation models to synthesize robot policies from language and visual demonstration,'' in \emph{Annual Conf. on Robot Learning}, 2024.

\bibitem{wake2024gpt}
N.~Wake, A.~Kanehira, K.~Sasabuchi, J.~Takamatsu, and K.~Ikeuchi, ``Gpt-4v (ision) for robotics: Multimodal task planning from human demonstration,'' \emph{IEEE Robotics and Automation Letters}, 2024.

\bibitem{wang2024large}
J.~Wang, Z.~Wu, Y.~Li, H.~Jiang, P.~Shu, E.~Shi, H.~Hu, C.~Ma, Y.~Liu, X.~Wang \emph{et~al.}, ``Large language models for robotics: Opportunities, challenges, and perspectives,'' \emph{arXiv preprint arXiv:2401.04334}, 2024.

\bibitem{sharma2022correcting}
P.~Sharma, B.~Sundaralingam, V.~Blukis, C.~Paxton, T.~Hermans, A.~Torralba, J.~Andreas, and D.~Fox, ``Correcting robot plans with natural language feedback,'' \emph{arXiv preprint arXiv:2204.05186}, 2022.

\bibitem{bucker2023latte}
A.~Bucker, L.~Figueredo, S.~Haddadin, A.~Kapoor, S.~Ma, S.~Vemprala, and R.~Bonatti, ``Latte: Language trajectory transformer,'' in \emph{2023 IEEE Intl. Conf. on Robotics and Automation (ICRA)}.\hskip 1em plus 0.5em minus 0.4em\relax IEEE, 2023, pp. 7287--7294.

\bibitem{shi2024yell}
L.~X. Shi, Z.~Hu, T.~Z. Zhao, A.~Sharma, K.~Pertsch, J.~Luo, S.~Levine, and C.~Finn, ``Yell at your robot: Improving on-the-fly from language corrections,'' \emph{arXiv preprint arXiv:2403.12910}, 2024.

\bibitem{yu2023language}
W.~Yu, N.~Gileadi, C.~Fu, S.~Kirmani, K.-H. Lee, M.~G. Arenas, H.-T.~L. Chiang, T.~Erez, L.~Hasenclever, J.~Humplik \emph{et~al.}, ``Language to rewards for robotic skill synthesis,'' \emph{arXiv preprint arXiv:2306.08647}, 2023.

\bibitem{cui2023no}
Y.~Cui, S.~Karamcheti, R.~Palleti, N.~Shivakumar, P.~Liang, and D.~Sadigh, ``No, to the right: Online language corrections for robotic manipulation via shared autonomy,'' in \emph{Proceedings of the 2023 ACM/IEEE Intl. Conf. on Human-Robot Interaction}, 2023, pp. 93--101.

\bibitem{wu2024general}
J.~Wu, Y.~Jiang, Q.~Liu, Z.~Yuan, X.~Bai, and S.~Bai, ``General object foundation model for images and videos at scale,'' in \emph{Proceedings of the IEEE/CVF Conf. on Computer Vision and Pattern Recognition}, 2024, pp. 3783--3795.

\bibitem{openai2024gpt4o}
\BIBentryALTinterwordspacing
OpenAI, ``Gpt-4o system card,'' 2024. [Online]. Available: \url{https://arxiv.org/abs/2410.21276}
\BIBentrySTDinterwordspacing

\bibitem{radford2023robust}
A.~Radford, J.~W. Kim, T.~Xu, G.~Brockman, C.~McLeavey, and I.~Sutskever, ``Robust speech recognition via large-scale weak supervision,'' in \emph{Intl. Conf. on machine learning}.\hskip 1em plus 0.5em minus 0.4em\relax PMLR, 2023, pp. 28\,492--28\,518.

\end{thebibliography}

\section*{Subjects' Requests}
\label{appendix}

\paragraph*{Subject 1}
\begin{itemize}
    \item I clean the tray with the sponge.
    \item After this movement, please clean also the bottom part of the tray. \\
\end{itemize}

\paragraph*{Subject 2}
\begin{itemize}
    \item From step 2 to step 5 you have to clean the tray, then add a rotation of minus 45 degrees and move to bottom left corner touching the tray, then rotate to 0 degree and move to bottom right corner touching the tray.
    \item From step 2 to step 5 you have to clean the tray.
    \item Modify the angle degrees from step 2 to 5 to 0 degrees.
    \item After step 5, continue to clean the tray to bottom left corner and then move to bottom right corner, always with 0°. \\
\end{itemize}

\paragraph*{Subject 3}
\begin{itemize}
    \item I want to clean the tray.
    \item To clean the whole tray, I want to pass by the bottom left corner and the bottom right corner after the right edge middle point. \\
\end{itemize}

\paragraph*{Subject 4 (naive)}
\begin{itemize}
    \item I cleaned the tray, but I want to add also bottom left corner and bottom right corner. \\
\end{itemize}

\paragraph*{Subject 5}
\begin{itemize}
    \item In my application, I want to clean all the surface of the tray with the sponge. The robot should touch top left corner and then go to top right corner while touching. And continue to go bottom left corner while touching. And then it should go bottom right corner again touching. \\
\end{itemize}

\paragraph*{Subject 6}
\begin{itemize}
    \item It's a tray cleaning task in the video. We have to touch the tray and we will start from the top right corner, top edge midpoint, top left corner, then right edge midpoint, center, left edge midpoint, bottom right corner, bottom edge midpoint and bottom left corner. That should be the order. \\
\end{itemize}

\paragraph*{Subject 7 (naive)}
\begin{itemize}
    \item From step 3 to step 5, you should be touching instead of above.
    \item Between step 3 and step 4 it should go to right edge midpoint.
    \item Step 5 should be bottom left corner.
    \item Add a final step to go to bottom right corner. \\
\end{itemize}

\paragraph*{Subject 8 (naive)}
\begin{itemize}
    \item Substitute from step 2 above with touching, and after go to bottom left corner and after bottom right corner.
    \item Substitute from step 2 above with touching. I want to complete the cleaning of the tray.
    \item I want to complete the cleaning of the tray. \\
\end{itemize}

\paragraph*{Subject 9 (naive)}
\begin{itemize}
    \item From step two, when you're on the top left corner, you start to keep touching the tray, and then you will go to top right corner with touching. And then in step four, instead of left edge mid, you will go to bottom left corner, and then you will go to bottom right corner, and it will be finished.
    \item When you go to the top right corner, you will wipe the middle side, which is from left edge midpoint to right edge midpoint, and then you go bottom left corner. \\
\end{itemize}

\paragraph*{Subject 10}
\begin{itemize}
    \item First of all, you're wrong. The sponge should touch the tray.
    \item Could you please add two steps reaching the bottom left corner and the bottom right corner?
    \item Could you please add at the end of the plan the reach of the bottom left corner?
    \item Could you please add as last step the reaching of the bottom right corner?
\end{itemize}  

\end{document}